# AUTOMATING CREATIVITY

Ming-Hui Huang[1*], Roland T. Rust[2]

March 14, 2024

[1*]Department of Information Management, College of Management, National Taiwan University; Taipei 10617, Taiwan.

[2]Department of Marketing, Robert H. Smith School of Business, University of Maryland; College Park, MD 20742, USA.

*Corresponding author. Email: huangmh@ntu.edu.tw

**Abstract:** Generative AI (GenAI) has spurred the expectation of being creative, due to its ability to generate content, yet so far, its creativity has somewhat disappointed, because it is trained using existing data following human intentions to generate outputs. The purpose of this paper is to explore what is required to evolve AI from generative to creative. Based on a reinforcement learning approach and building upon various research streams of computational creativity, we develop a triple prompt-response-reward engineering framework to develop the creative capability of GenAI. This framework consists of three components: 1) a prompt model for expected creativity by developing discriminative prompts that are objectively, individually, or socially novel, 2) a response model for observed creativity by generating surprising outputs that are incrementally, disruptively, or radically innovative, and 3) a reward model for improving creativity over time by incorporating feedback from the AI, the creator/manager, and/or the customers. Two proofs of concept are illustrated by developing Starbucks' and iPhone's marketing strategies following the framework. We derive conceptual implications from the framework for making GenAI creative, such as maximizing the novelty of prompts objectively and subjectively, exploring the unexpected responses by showing what is expected, and rewarding prompt-response pairs based on strategic goals. We further derive managerial implications from the framework, such as leveraging the data and search capabilities of GenAI for incremental innovation (e.g., cell phones adding new features), exploring the boundary of expectation to achieve disruptive innovation (e.g., iPhone competes in the cellphone market with a set of smart features), and reversing interaction using two-way brainstorming for radical innovation (e.g., iPhone creates the smartphone market). This framework enables the application of GenAI for various levels of creativity strategically.

**Acknowledgements**
This research was supported by grants (MOST 110-2410-H-002-101-MY3 and NSTC 112-2410-H-002-048-MY3) from the National Science and Technology Council, Taiwan.

# AUTOMATING CREATIVITY

## Introduction

Generative AI (GenAI)[1] is the class of AI systems that can generate outputs, which contrasts with its previous generation of discriminative AI that is mainly for recognizing inputs (e.g., content classification, image recognition). OpenAI's GPT models (e.g., ChatGPT or GPT-4) are state-of-the-art GenAI that have received wide attention since their launches in late 2022 and early 2023. They also spur the global GenAI market size, which Acumen Research and Consulting (December 2022) has forecasted to reach $110 billion by 2030.

GenAI, such as GPT, is an autoregressive model that is trained using both public data and third-party licensed data to predict the next word for a given input (typically textual—for more details on the algorithm, see Radford et al., 2019) following a prompt-response paradigm with user[2] input as "prompt" instructing GenAI what to generate, and with GenAI output as "response." The capability of GenAI to generate outputs has increasingly jeopardized human creativity, as these tasks and jobs do not seem to be safe from AI replacement (Huang and Rust 2018).

The creative capability of GenAI has received media and research attention (Epstein et al., 2023; Iyengar, 2023). However, it has often been noted that the outputs generated by GenAI have limited creativity and originality, because the outputs are generated based on learning from existing data, and GenAI is pre-trained to maximize the likelihood of its training data; thus, it is difficult for GenAI to go beyond the training data and think outside the box (Korbak et al., 2023;

---
[1] Franceschelli and Musolesi (2022) survey seven types of creative generative models, encompassing variational autoencoders (VAEs), generative adversarial networks (GANs), sequence prediction models (CNN, RNN, LSTM), transformer-based models (including GPT, ChatGPT), diffusion models, reinforcement learning models, and models that depend on input. These models differ in their fundamental concepts and optimization techniques to foster creativity, indicating that not all generative models rely on transformer architecture. Our framework zeroes in on transformer-based generative models viewed through the lens of reinforcement learning. Transformer-based generative models are attributed by many, including Bommasani et al. (2021) and Franceschelli and Musolesi (2022), as the foundation model of GenAI.

[2] The "user" can refer to the developer, the creator, the manager, or the customer who interacts with GenAI for creativity.

Ouyang et al., 2022; Wong et al., 2023). Rust and Huang (2021) illustrate the current limitations of AI creativity using the examples of an AI-generated impressionist style painting from *Bloomberg Businessweek* (May 17, 2018) and a Lexus car TV commercial. By feeding machines with tones of impressionist painting, AI combined components in these existing paintings into a new painting. The Lexus car advertisement script was generated by learning award-winning luxury car commercials, Lexus brand information, and emotional data. This approach was used to narratively present the new Lexus ES executive saloon car. The painting and the commercial both generated wide attention, but the degree to which they go beyond the data is questioned. More recently, using GenAI in home renovation, Patrick Murphy, Cofounder and CEO of Market.AI, states that lack of creativity is one of the biggest limitations of GenAI. He puts it this way, "Generative AI is only as good as the data it's been trained on, so it can't come up with totally original designs or come up with new solutions to problems" (March 14, 2023).

Academic research echoes a similar concern. Bubeck et al. (2023) state that GPT-4 is not particularly effective at generating content that requires the identification or creation of novel perspectives or frameworks to address a problem, for example, writing a joke, coming up with a scientific hypothesis, or creating a new style of writing. Wong et al. (2023) argue that the techniques used by GenAI to create output in response to prompt (e.g., fine tuning, prompt tuning, prompt engineering, or in-context learning) often generate a single output that satisfies user preference to some extent, but often involves a tradeoff between remaining faithful to the prompt and generating diverse outputs. Thus, the way that GenAI is trained, although enabling it to combine existing ideas into new outputs, is the lowest level of creativity (López de Mántaras, 2016; Boden, 2004). The AI creativity pioneer Boden (1998) even says that "Creativity is a fundamental feature of human intelligence, and a challenge for AI."

Because GenAI currently is not considered to be very creative, we propose a prompt-response-reward engineering framework to evolve it from generative to creative. The framework comprises three models: 1) prompt engineering for creative prompt development, 2) response engineering for creativity generation, and 3) reward engineering for creativity evaluation and improvement. The proposed framework has the potential to make AI creative, considered by many as a big step toward artificial general intelligence (AGI).

Our paper contributes to information systems (IS) in multiple aspects. First, at the organizational level, it has implications for organizational innovation. Creativity is important for organizational innovation (Couger et al., 1993; Massetti, 1996; Seidel et al., 2010), which is an important organizational resource for competition. For example, the capability to develop the ground-breaking innovative iPhone made Apple Computer highly competitive in the smartphone market. Amabile (2020) uses the two terms, creativity and innovation, side-by-side to show the important role of AI in organizational creativity and innovation. The proposed framework provides guidelines for companies to use GenAI for various levels of innovation strategically.

Second, at the individual level, creativity is an important human skill, but not everyone has the talent to be creative. GenAI can be creative on its own or can serve as human creativity support. Some previous information systems have served this purpose but at a very limited level. State-of-the-art GenAI has demonstrated great potential for this purpose. Many marketing specialists have used GenAI to help develop search ads, social media ads, and many other content generations that improve marketer creativity. The proposed framework provides a stagewise process for individuals to use GenAI for creativity.

Third, the proposed framework has implications for IS researchers and practitioners about how to evolve AI from generative to creative. This is important for the development of AI, since so far, we don't have AGI yet, and state-of-the-art GenAI is not very creative. Many of the concepts discussed in the prompt-response-reward framework suggest directions for what needs to happen for GenAI to be creative. This is an important task for IS scholars.

Fourth, the triple engineering framework provides a new perspective to both IT innovation (i.e., process innovation) (Massetti, 1996; Cooper, 2000; Couger et al., 1993; He et al., 2022; Wang, 2021) and digital product innovation (i.e., output innovation) (Lehmann et al., 2022; Wang et al., 2022). The triple engineering framework details the process of how GenAI, as a generative technology, can enhance its creativity by prompts, responses, and rewards, and the outcome is service innovation (digital products are service) that varies in levels of creativity.

## Computational Creativity Defined

### Creative AI

In the computing literature, computational creativity is generally viewed as the processes or activities for computational systems to generate outputs that are evaluated by those systems or human observers as creative (Franceschelli & Musolesi, 2022; Lamb et al., 2018; Mateja & Heinzl, 2021; Wiggins, 2006; Wu & Miao, 2013; Xiao et al., 2019). In the context of this paper, we specifically refer to GenAI as the computational system.

This definition of computational creativity indicates that for AI to be creative, the AI system needs to be able to process inputs to generate outputs that are deemed (by the AI system or humans) as creative. Thus, three components (stages) are involved in AI creativity: input processing, output generation, and output evaluation. Inputs can be external stimuli in the environment and the internal provided (or trained) data (Mateja & Heinzl, 2021), outputs can be intangible ideas such as jokes, musical compositions, scientific theories, marketing strategies, problem solutions, or digital products (Lehmann et al., 2022; Wang et al., 2022); or tangible artifacts such as paintings, poems, and music (Boden, 1998), and the evaluation can be carried out by the AI systems or by human observers (Franceschelli & Musolesi, 2022).

### Creative Concept

Boden (1990) defines a concept as an abstract idea in arts, science, and everyday life. Xiao et al. (2019) refer to "concept" as a bewildering range of things. Basically, these views show that creative concepts can be anything, if they are novel, surprising, and valuable (Alhussain & Azmi, 2021; Boden, 1998, 2004). The three criteria define whether a concept is creative. Among the three criteria, novelty and value are the two that are more consistently recognized as the essential criteria for creativity (Ghosh & Wu, 2021; Rosengren et al., 2020). For example, He et al. (2022)

consider creativity as an individual's ability to generate new ideas and useful ideas. Massetti (1996) uses the generation of novel ideas and valuable ideas as the two benchmarks for gauging whether using a creativity support system enhances individual creativity.

The three criteria are not linear compensatory—a concept needs to meet all three criteria to be considered as creative. They also are relative, meaning that depending on which domain a concept belongs to and who evaluates the concept, the result will be different. Table 1 summarizes the three creativity criteria, keywords, and relevant literature. We discuss the three creative concept criteria individually below.

**Table 1. Creativity Criteria, Keywords, and Literature**

| Criterion | Definition | Keywords | Literature |
| --- | --- | --- | --- |
| Novelty | A novelty concept is new and dissimilar with respect to existing concepts in a conceptual space | novel, new, original, unusual, distinctive, dissimilar, different, untypical, unknown, rare | Alhussain and Azmi (2021); Amabile (2020); Bartel (1985); Constantin et al. (2020); Boden (1998; 2004); Massetti (1996); Ghosh and Wu (2021); He et al. (2022); Hunt and Worthen (2006); Im and Workman (2004); Maher (2010); Wu and Miao (2013) |
| Surprise | A surprising concept is unexpected, unfamiliar, and impossible to the user | surprising, unexpected, unfamiliar, impossible, unlikely | Boden (1998, 2004); Constantin et al. (2020); Lamb et al. (2018); Wu and Miao (2013) |
| Value | A valuable concept is a high-quality instance in its class and has high utility preferences to the target customers (e.g., interesting, useful, beautiful, attractive, etc.) | high quality, utility preference, aesthetic, interesting, useful, beautiful, attractive, appropriate, correct, meaningful, relevant, affective, pleasure, appropriateness | Alhussain and Azmi (2021); Amabile (2020); Boden (1998, 2004); Massetti (1996); Constantin et al. (2020); Ghosh and Wu (2021); He et al. (2022); Im and Workman (2004; Rosengren et al. (2020); Wu and Miao (2013) |

**Note.** The "user" can include the developer, the creator, the manager, and the customer, while the "target customers" are more specific about the individuals who consume the creativity.

**Novelty**

A novel concept is new, original, dissimilar (different), unique, untypical, unusual, and distinct with respect to existing concepts in a conceptual space (Ritchie, 2007). A conceptual space encompasses a universe of all entities (artifacts, concepts, ideas) that are producible following certain rules (Lamb et al., 2018), a style of thinking; for example, a way of writing poetry, a style of music, or a theoretical area in marketing (Boden, 2009), a set of concepts that are in some sense deemed to be acceptable as examples of whatever is being created (Wiggins, 2006), a class that an artifact belongs (Constantin et al., 2019), or a product market (Ghosh & Wu, 2021).

Novelty is benchmarked by the existing concepts in the conceptual space where the new concept belongs. It can be "original" (Bartel, 1985; Constantin et al., 2019), i.e., the first one in its class; or "dissimilar", i.e., different from other concepts in the conceptual space (Hunt & Worthen, 2006; Ritchie, 2007). Novelty makes the concept or artifact easily distinguishable (different) from all others of the same class (Hunt & Worthen, 2006).

To summarize, a novel concept is dissimilar from existing concepts in a conceptual space, e.g., in the business world, a brand (a concept) is competing in a market (a conceptual space). This criterion is objective to machines, but is subjective to humans.

**Surprise**

Surprise means that the concept is unexpected, unfamiliar, or unlikely to occur to the evaluator (Boden, 1998, 2004), where the evaluator can be the machine (i.e., GenAI itself), the creator (e.g., an artist, a developer, or a manager) or the customers (who consume the new concept or artifact, e.g., buyers of the artist's painting, users of a creative system, or customers of a new product). Compared to the "novelty" criterion, the "surprise" criterion shifts the benchmark from the existing concepts in the conceptual space to the evaluator's expectation. Expectations are developed from experience. Surprise arises when there is an expectation, and the concept contradicts the expectation (Constantin et al., 2019; Roseman, 1996; Wu & Miao, 2013). Lamb et al. (2018) consider that the more unexpected the new patterns are compared to prior frameworks, the more creative they are considered, reflecting the surprise criterion of creativity. Cooper (2000) argues that previous experience could lead to a functional fixedness that reduces the likelihood of producing creative solutions, suggesting that meeting expectations is a less creative solution. Lehmann et al. (2022) argue that the generative potential of digital technology (e.g., artificial intelligence) can offer novel market offerings in a way that is not intended or imagined by the inventor.

To summarize, a surprising concept is one that is unexpected to the evaluator. The evaluator can be GenAI that generates the concept, the creator of the concept, or the consumers who appreciate the concept. In the business world, the creator can be the IT manager and the users are the customers of the IT solution. This criterion is subjective, reflecting the experience or expectation of the creator and/or the customers.

**Value**

A valuable concept is a high-quality instance in the conceptual space, and is preferred by the evaluator, for example, useful (He et al., 2022; Ghosh & Wu, 2021), meaningful (Massetti, 1996); appropriate (Rosengren et al., 2020), or interesting, beautiful, and attractive (Constantin et al., 2019). Useful, meaningful, and appropriate reflect the quality aspect of a concept in relative to other concepts, and interesting, beautiful, and attractive reflect the consumer preference aspect of the concept.

The two aspects of the criterion reveal that value is benchmarked against both the existing concepts in the conceptual space and the evaluator's preference. In Ritchie's Criteria (2007), valuable is evaluated by quality, which measures how much a creative output is a high-quality example of its genre.

Value also needs to have utility preferences to users, for example, an application appeals to the intended market (Ghosh & Wu, 2021), a creative development task is satisfying to an individual in an organization (Cooper, 2000), a digital technology enhances user experience (Wang, 2021), or a concept is seen as "beautiful" or appealing (Weisberg, 1999), and elicits satisfaction, attraction, and emotional well-being from the user (Constantin et al., 2019; Zangwill, 2003).

Preferences vary across different individuals and different cultures. Constantin et al. (2019) state that conventional wisdom holds that "beauty is in the eye of the beholder," implying that aesthetic judgments are subjective and influenced by the individual characteristics of the observer. This subjectivity results in the need for evaluation by the customers (Huang, 2023), because value cannot be gauged by a set of objective metrics only (unless the set of objective metrics have been paired with the consumer's preferences).

To summarize, a valuable concept is a high-quality instance of its class, is unexpected to the evaluator, or elicits high utility preferences from the consumers.

### *Creativity Level*

Boden (1998, 2014) distinguishes three levels of computational creativity, based on varying degrees of surprise the creativity generates: combinational, exploratory, and transformational. Combinational creativity generates the lowest degree of surprise, which is a new combination of existing concepts in the existing conceptual space, such as the responses that ChatGPT can generate from the "summarization" category of prompts. Exploratory creativity goes one level up with new concepts in an existing conceptual space being generated. The "generation" category of prompts has a higher probability to achieve this level. Transformational creativity has the greatest surprise, with new concepts being generated that define a new conceptual space such as what the "brainstorming" category of prompts may be able to generate. Kirkpatrick (2023) provides a plain language interpretation of the three creativity levels: 1) generating new ideas from combinations of familiar ideas, 2) generating new ideas based on the characteristics of existing works, and 3) generating new ideas from combination of characteristics and ideas that were previously unconsidered by humans in the creation process.

**Combinational Creativity**

This is the lowest level of creativity, because it does not go beyond the data (conceptual space) used to pre-train GenAI; thus, it generates the lowest degree of surprise. Take, for instance, a brand-new rock 'n' roll tune that merges elements from different musical styles. Even though it introduces some innovative features, the unmistakable essence of rock 'n' roll is immediately recognizable to listeners (Rust & Huang, 2021). All the pre-trained models share this limitation. What they can generate is limited by the data used to train them. Boden (2004) states that combinational creativity is easy for a computer to do, because there is nothing more straightforward than selecting two ideas and placing them side by side. Andrus (1968) discusses computer creativity in marketing as mechanical creativity that computers routinely combine and sort stated elements creatively. Lehmann et al. (2022) discuss that generative digital technology (e.g., artificial intelligence) can generate novel market offerings by combining digital technology elements in a way that was not intended or imagined by the inventor.

Combinational creativity is like incremental innovation that makes continuous improvements to existing products without changing the fundamental design; for example, the regular updates to iPhone by improving the camera quality, increasing battery life, or enhancing the processing speed in the newer models.

**Exploratory Creativity**

Exploratory creativity is more creative than the combinational version and generates a higher degree of surprise. New concepts are generated by exploring the boundary of an existing conceptual space. Lamb et al. (2018) state that exploratory creativity, through experimenting with the implications of rules, reaches points within the space that were previously have not been reached before. Bob Dylan, for example, fused the protest songs of Woody with the energy of rock 'n' roll, crafting a novel style characterized by both a compelling rhythm and lyrical depth. This innovation had a profound impact on major 1960s musicians, including Jimi Hendrix and the Beatles. In recognition of his influence, Dylan was awarded the Nobel Prize in Literature (Rust & Huang, 2021). Such exploratory creativity often requires a change of perspective to expose the boundary of the existing conceptual space, or to have a new look at the existing space. In theorizing digital innovation ecosystems, Wang (2021) considers that a digital innovation ecosystem facilitates innovations because it can bring in external knowledge for solving a particular problem (e.g., enable the firm to solve the problem from a different perspective). He illustrates a successful case of Apple Computer. By orchestrating its ecosystem, the company generates an unprecedented wealth of innovations.

Exploratory creativity is like disruptive innovation, such as a new technology/feature that disrupts the existing market; for example, the launch of the iPhone in 2007 disrupted the mobile phone industry by integrating an internet browser, music player, touch screen interface, and phone into a single device that set a new standard for smartphones. Lehmann et al. (2022) argue that there is a tension between new market offerings generated by generative digital technologies and the existing environment (i.e., the prevalent industry structure). When the new offerings are incompatible with the existing market, it can backfire (e.g., leave customers and complementors overwhelmed) even though it might be successful in the future. This shows the dilemma about how far to go for the creativity level—the more creative, the longer the return term, the higher risk, but also possibly a higher return.

**Transformational Creativity**

Transformational creativity is the highest level of creativity. It is rare, and generates the highest degree of surprise. This level of creativity brings forth novel concepts that establish a new conceptual space. For example, in the 1970s, the German band Kraftwerk introduced a groundbreaking performance style by appearing on stage exclusively with synthesizers and drum machines, some of which were their own inventions. This marked a significant divergence from historical musical norms, involving not just new instruments but also a fresh type of music that paved the way for the emergence of modern electronica and even hip hop. Transformational creativity redefines the rules to reach points that were once beyond reach. The more inconceivable the new patterns are within old frameworks, the higher level of their creativity (Boden, 2004; Lamb et al., 2018). For GenAI, this may require re-training using brand new data so that new ideas can be generated from a new conceptual space. Re-training requires large-scale data, which would be difficult to achieve in many cases.

Transformational creativity is like radical innovation, acting as a technological revolution that transforms industries and establishes new market landscapes. The introduction of the iPhone app store in 2008 created a platform where developers could create and sell applications directly to users, changing the way mobile software was distributed, but also enabling a plethora of new businesses and industries to emerge. Apple computer's graphical user interface represents another example of radical innovation such that the new components of icons, mice, windows, new relationships among those components, and new relationship with users radically departure from the prevailing interfaces at that time (Cooper, 2000). ChatGPT is a recent example of radical innovation. By designing it as an interactive system between AI and users, it radically transforms how users interact with AI.

## The Prompt-Response-Reward Engineering Framework for AI Creativity

We develop a triple prompt-response-reward engineering framework to evolve AI from generative to creative. Figure 1 illustrates this triple engineering process from a reinforcement learning approach. Bold typeface denotes the terms in reinforcement learning, while regular typeface denotes the corresponding terms in the triple engineering framework. Figure 2 illustrates the dynamic of the three models. Table 2 presents the components of the framework for creativity and its implications.

## Table 2. The Prompt-Response-Reward Framework for Creativity and Its Implications

| The Model | Components | | | Implications |
|---|---|---|---|---|
| **Model 1 Prompt engineering** (creative input identification) Engineer novel prompt inputs for expected creativity | \<center\>**Novelty**\</center\> | | | |
| | Specify prompt space | Decide prompt value function (Define novelty) | Engineer prompt value | • Specify the product market as the boundary of the prompt space |
| | • The richer, more diversified, better linked, better sequenced, and more effective the space, the more likely for a prompt from this space to generate creative outputs | • Assign expected novelty value to prompts by observing objective novelty, individual novelty, and social novelty | • Develop novel prompts for generating creative responses | • Identify the target customers to decide the subjective novelty<br>• Generate new products/features that are novel to the market and can be expected to be preferred by the target customers |
| **Model 2 Response engineering** (creative output generation) Engineer surprising responses for observed creativity | \<center\>**Surprise**\</center\> | | | |
| | Search response space | Decide response value function (Define expectation) | Engineer response strategy | • Search product/feature space for unfamiliar combinations (incremental innovation) |
| | • Search for <u>unfamiliar</u> response combinations as new responses | • Strategy demonstration and flexible strategy search to <u>go beyond expectations</u> | • Reverse interaction (two-way brainstorming) to explore <u>improbability</u> | • Explore the boundary of expectation (disruptive innovation)<br>• Reverse interaction to generate new products that define a new market (radical innovation) |
| **Model 3 Reward engineering** (creative output evaluation) Engineer prompt-response evaluations for better creativity | \<center\>**Value**\</center\> | | | |
| | Identify reward space | Design reward function (Define creativity/strategic goal) | Engineer rewards | • Prioritize creativity rewards<br>• Make creativity strategic<br>• Balance the sources of feedback for improvement |
| | • Novelty<br>• Surprise<br>• Value | • New segment to target (discover new segment)<br>• Innovative product design for the segment (product innovation)<br>• Benefit to the segment (customer value) | • Learn from AI feedback<br>• Learn from expert feedback<br>• Learn from customer feedback | |

**Note.** The concept of "target customers" refers to aggregate customers that represent a segment of the market.

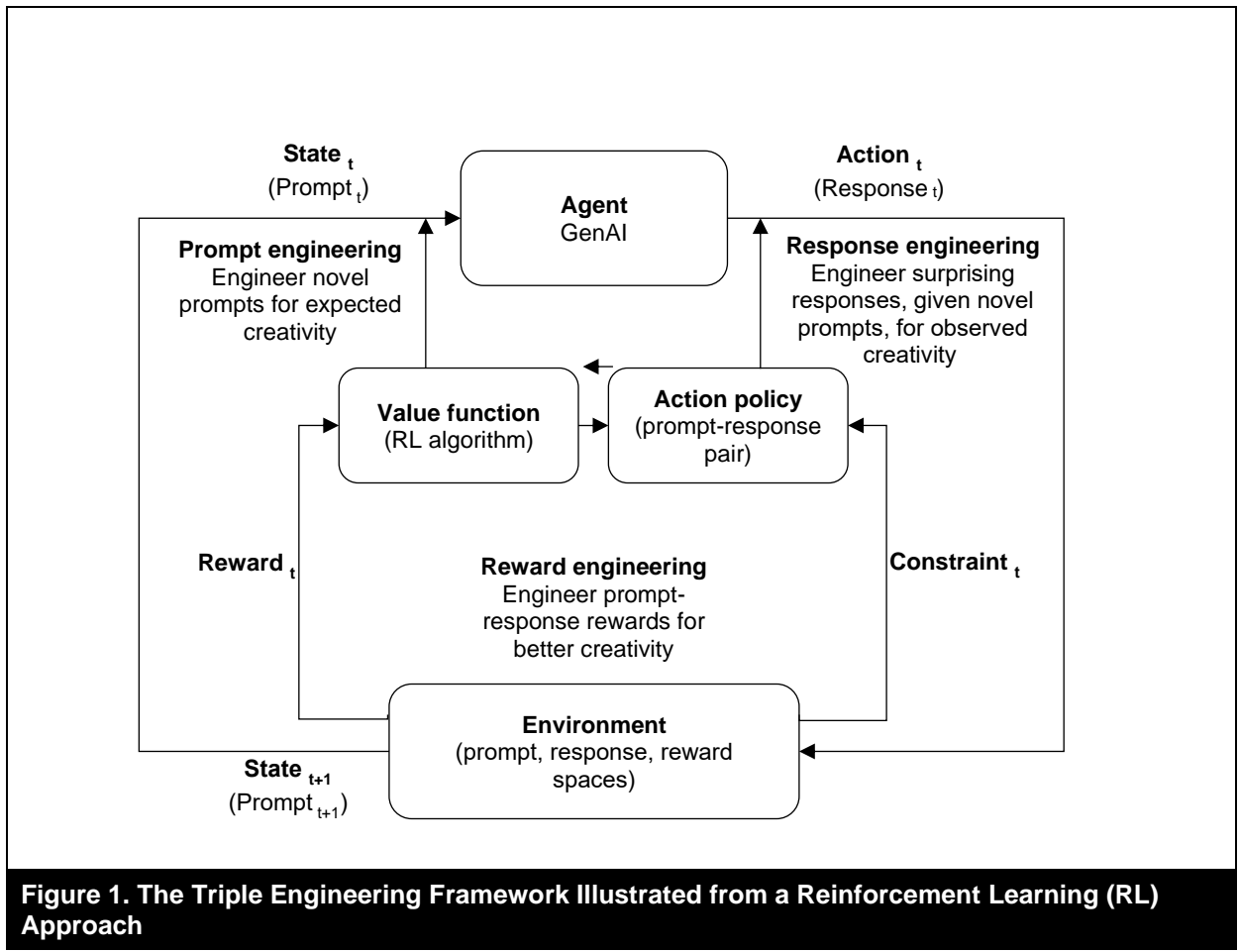

**Figure 1. The Triple Engineering Framework Illustrated from a Reinforcement Learning (RL) Approach**

**Note.** Bold typeface denotes terms in reinforcement learning, while regular typeface denotes the corresponding terms in the triple engineering framework. Action policy can be pre-trained to become part of the agent (e.g., train GenAI to be helpful and harmless) or fine-tuned (as shown in the figure) with expert domain knowledge. The value function is updated by creator/manager evaluation. customer preferences, and/or machine self-evaluation. The environment is a conceptual space, which is populated by prompts, responses, and rewards. The "value function → policy" relationship denotes that the value function decides the value of the prompt-response pair, whereas the "value function ← policy" relationship denotes that the value function is conditional on the policy. The "environment → policy" relationship denotes that possible policies are constrained by the environment.

In reinforcement learning, an AI agent acts in the environment to maximize the rewards over time. The environment is populated by acceptable states, actions, and rewards. Given the current state, the agent decides an action based on the action policy, i.e., the strategy that dictates the action the AI agent takes at a given time. If the state-action pair is positive, the environment sends out a reward signal. The reward will update the value of the state and the subsequent state-action pair. The reward function defines the objective and the feedback for the agent so that the agent can take actions that are aligned with the objective (Cao et al., 2023; Franceschelli & Musolesi, 2022; Russell & Norvig, 2010; Sutton & Barto, 2018).

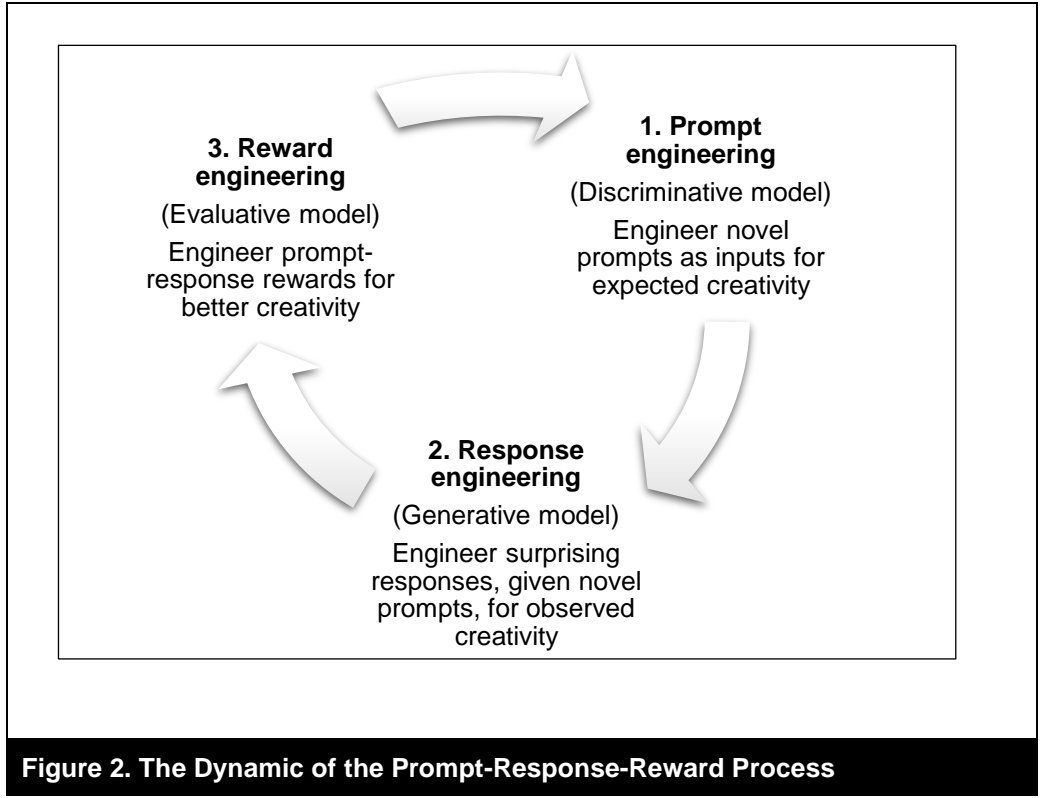

**Figure 2. The Dynamic of the Prompt-Response-Reward Process**

In a prompt-response paradigm, prompts are the human side of the conversations (Bai et al., 2022), responses are the GenAI side of the conversations, and rewards are the incentives for creativity improvement. In the triple engineering framework, the prompt model develops input prompts that are objectively novel, individually novel, or socially novel for expected creativity. The response model generates surprising creativity output responses that are unfamiliar new combinations of existing concepts (incremental innovation), unexpected to the manager (disruptive innovation), or deemed unlikely for the market (radical innovation). The reward model designs rewards to motivate and evaluate creativity for improvement, by learning from AI feedback (intrinsic reward), learning from creator/manager feedback (extrinsic reward), or learning from customer feedback (extrinsic reward). The three components together constitute a dynamic loop for identifying novel input prompts, generating surprising creative responses, and rewarding valuable creativity.

*Prompt Engineering: Prompt Development Model*

The prompt model is to develop discriminative prompts as novel inputs that are expected to generate creative outputs. The prompt inputs can be objectively novel to the machine, individually novel to the creator/manager, or socially novel to the market. Figure 3 illustrates the three type of novelty.

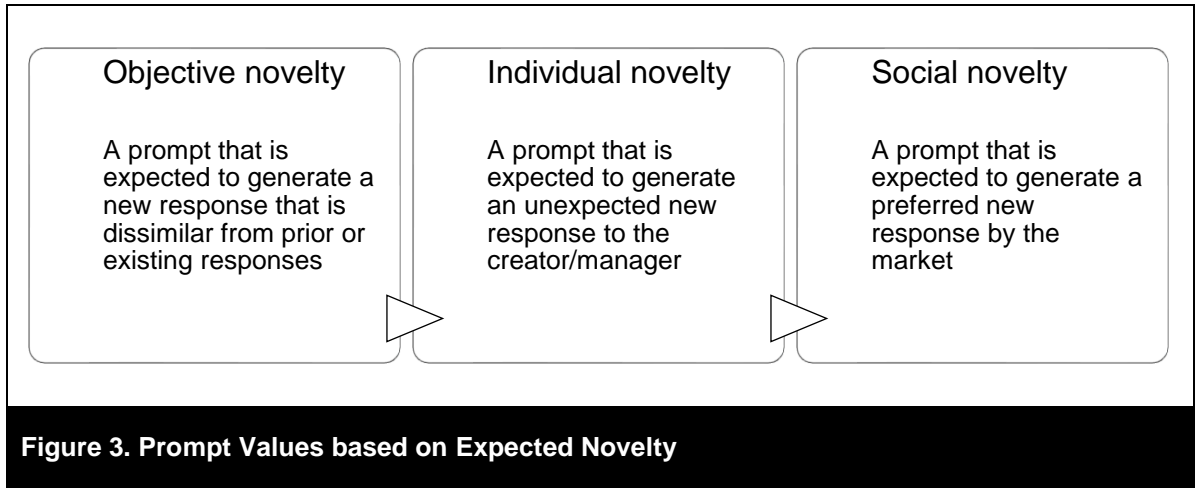

**Figure 3. Prompt Values based on Expected Novelty**

Prompt engineering, also referred to as "prompt design," "prompt programming," and "prompting," is how GenAI is programmed via prompts to solve the task at hand (Liu et al., 2023; Oppenlaender et al., 2023; White et al., 2023). A prompt is an instruction or a set of instructions, usually in free-form natural language (spoken or written) but can also be images, providing rules and guidelines telling GenAI what the desired output should be (Bach et al., 2022; Liu et al., 2023; White et al., 2023; Wong et al., 2023). In prompt engineering, the objective is to develop a prompting function, $y = f_{prompt}(x)$, designed to yield the most effective output $y$ from the given input $x$ (Liu et al., 2023).

Prompt engineering can be automated or manually done. In automated prompting, the machine should be designed as a discriminative model to identify the most novel prompts that have the highest information value for GenAI to generate creative outputs, which can be achieved by using prompt templates (Liu et al., 2023). In manual prompting, this process is for the creator to develop prompts and GenAI to generate responses. An effective prompt establishes the context of the interaction, informs GenAI about the crucial information, and specifies the expected format and content of the output (White et al., 2023). The prompt capable of generating the desired response is often refined over several iterations to enhance its effectiveness.

Below we discuss the three steps of prompt engineering: first, specify the prompt space to tell GenAI the boundary of knowledge search or retrieval; second, define the prompt value function to decide the value of a prompt; and third, engineer the prompt value to develop novel prompt for expected creative response.

**Specify the Prompt Space**

The prompt space is a collection of possible prompts that GenAI can take as inputs to generate the creative outputs. The task of prompt engineering is to identify prompts from the space that can make GenAI generate creative outputs.

In computational creativity, a conceptual space is where all acceptable creative concepts, current and future, are located (Wiggins, 2006). They are all things that can be generated according to a set of rules (Lamb et al., 2018). The goal is to select data to enable machine creativity (Mateja & Heinzl, 2021). Bai et al. (2023) consider this process as feature discovery, in which the agent adds features to the data as inputs for better learning. The space can be specified either declaratively or procedurally (Xiao et al., 2019). However, in the era of big data, the quantity of data requiring processing and learning span multiple domains. It's recognized that not all data points are equally significant for learning. Therefore, the focus should be on selecting the most relevant data and uncovering meaningful patterns, which is essential for reducing computational expenses and enhancing the efficiency of data usage (Sun et al., 2022).

In the prompt-response paradigm, a knowledge domain can be specified for developing the prompts, for example, by providing context for GenAI to retrieve relevant knowledge for response. A clear domain helps to draw a boundary between what is relevant and irrelevant. Specifying the prompt space also directs GenAI's attention to the under-explored prompt space, from which relevant feature representation can be extracted (Sun et al., 2022). In general, within the specified prompt space, the richer and the more diversified the knowledge (LeCun, 2022; Yao et al., 2023), the better the ability to form links of many different types of knowledge (Boden 2004), the better sequenced (e.g., state, action, and resulting state) (LeCun, 2022), and the more effective the space (Bai et al., 2023), the more likely for the AI to be creative.

In marketing, a prompt space can be seen as a product market where a brand is competing against other brands. For example, a sedan is competing in an automobile market, with the automobile market constituting the prompt space and with all sedans in the market populating the space. The boundary of the market is fuzzy. Considering that a consumer not only can consider different brands in the automobile market, but they can also consider the metro as an outside the box option, which is in a different or broader market such as transportation. Thus, prompt engineering needs to specify the prompt space so that prompts can be developed from an appropriate and relevant space to generate creative outputs.

**Decide Prompt Value Function**

The prompt value function, together with the response value function (to be discussed in the response model), comprise the two-part value function. Prompt value is the expected value. After GenAI generates a response based on the prompt, the observed value of the prompt can be decided.

Among the three creativity criteria, prompt value hinges more on novelty. Novelty is more about the input, whereas surprise is more about the output. For example, Bai et al. (2023) define a novel search as an agent's "behaviors" being different from previous behaviors, and define a surprising search as the "solution's" deviation from prior solutions. Wu and Miao (2013) discuss computational curiosity and consider detecting and adapting to novel inputs to be an important requirement for AI to have human-like curiosity. Novelty is contingent on time and a comparison of the present state with past experiences.

The prompt value function assigns the value to prompts based on the expected novelty. The expected novelty of a prompt can be evaluated by whether the prompt can generate a new response that is dissimilar from prior or existing responses (objectively novel), can generate an unexpected new response to the creator/manager (individually novel), or can generate a preferred new response by the market (socially novel).

To evaluate novelty is sometimes somewhat objective, for example, by calculating the distance between the concept and other concepts in the space. Deciding subjective novelty, by contrast, requires specifying whose expectations and preferences to consider. It can be the creator's/manager's expectations or the target customers' preferences (e.g., appreciators or buyers of the creativity), with the former being the subjective perception of an individual and with the latter being the shared experience in a culture (Berlyne, 1960).

In application, role specification is one such prompting method to ask GenAI to take perspective. The creator can ask GenAI to assume the role of a marketing specialist to generate a novel positioning slogan, or the creator can provide a definition of the target customers as part of the prompt to instruct GenAI to generate outputs that appeal to the target consumers, not the average consumers. White et al. (2023) provide a "persona pattern" prompting method. In this case, the creator asks GenAI's outputs to take a certain point of view or perspective. Contextual statements such as "act as persona X" or "provide outputs that persona X would create" are some examples.

**Engineer Prompt Value**

Prompt development should be based on discriminative models to identify novel representations of concepts that have the expected value of creativity. Wiggins (2006) defines representations as expressing concepts (artifacts) and rules. GenAI needs to identify prompts from the prompt space that have the potential to generate creative output. Mateja and Heinzl (2021) summarize four strengths of machine learning for enabling creativity: task finding, task preparation, response generation, and response validation. Engineering prompt value is a combination of the task finding and task preparation stages.

The learning for AI to be creative should be unsupervised. Kirkpatrick (2023) provides a detailed description about how this process works. GenAI is trained with vast arrays of examples across different formats, including images, audio clips, and pieces of text. It employs sophisticated pattern recognition techniques to identify unique features, patterns, or signals. For example, in creating art in the impressionist style, GenAI would examine works by renowned artists such as Monet, Renoir, Manet, Degas, Cezanne, and Matisse, recognized as giants in this field. It analyzes these artworks in terms of pixel patterns, identifying the unique patterns that characterize the impressionist style. This analysis forms a knowledge base that enables the generation of new works adhering to the learned parameters and attributes. The model's level of complexity and depth directly influences its ability to generate nuanced patterns and relationships. The effectiveness and quality of a GenAI model's outputs are directly linked to the quality of the datasets or training materials it has been fed.

In automated prompting, this process is to train GenAI to create and refine prompts itself to get the expected responses. This is considered as information extraction, such that structured information is extracted from unstructured prompts; for example, named entity recognition, relation classification, relation extraction, event detection, and event extraction (Li et al., 2023). In the process, the inputs are the unstructured prompts, which instruct GenAI to identify concepts (e.g., entities or relations) as prompts to generate creative outputs. They demonstrate that ChatGPT performs well in a setting where it must extract information openly, without a pre-defined set of labels, depending solely on its skill in understanding the prompt— this includes interpreting the task description, the question asked, and any provided input text—to produce predictions.

In manual prompting, the user needs to know how to ask questions or how to provide instructions for specific tasks. Manual prompting often relies on human intuition (i.e., human creativity) that is difficult to replicate. Thus, prompting skills are often deemed an art, rather than a science. White et al. (2023) argue that the quality of outputs from a conversational large language model (LLM) directly correlates with the quality of user-provided prompts. For example, Borne (2023) summarizes a short list of useful GenAI prompts that can generate creative outputs, such as creating content on any subject, writing cold emails and personalizing marketing copy to targeted customers, and letting GenAI become your mentor to guide and instruct you in learning new things.

### *Response Engineering: Response Generation Model*

The response model is to manage the responses generated by AI, given the prompts, to maximize the value of the prompt-response pair. This is the generation stage of AI creativity. Creative responses can be generated from re-using existing ideas (combinational), exploratory search for new ideas (exploratory), or transformational search beyond the conceptual space (transformational) (Boden, 2004). In reinforcement learning, the agent uses a policy to decide what action to take in a given state to maximize rewards over time. The policy is a probability distribution of actions for all states. It can be considered as a strategy that guides the actions. Response engineering focuses on developing strategies to generate the most valuable prompt-response pair.

We summarize from the literature and propose three strategies that can be used to engineer GenAI responses for the three levels of creativity: 1) response space search to generate unfamiliar responses (combinational creativity), 2) tree-of-thought demonstration to generate unexpected responses (exploratory creativity), and 3) reverse interaction to explore unlikely responses iteratively (two-way brainstorming) (transformational creativity). The three creativity strategies correspond to the three steps of response engineering: first, search the response space for unfamiliar responses; second, decide the value of the generated responses; and third, engineer a response strategy to maximize the value of the prompt-response pair.

**Search Response Space**

This is to tell GenAI where the boundary is for generating responses. The responses generated can be within the boundary (e.g., combinational creativity and exploratory creativity) or beyond the boundary (transformational creativity). It is like creating a smaller box (e.g., the response space for a specific task) within a bigger box (i.e., the conceptual space with the pre-training data as the boundary) so that responses can be generated at different levels of creativity. LeCun (2022) illustrates using an example to show that an agent's action typically impacts only a limited section of the space. For example, when an agent relocates a bottle from the kitchen to the dining room, it alters the states of the bottle, kitchen, and dining room, while the remainder of the conceptual space stays unchanged. Lehmann et al. (2022) describe a designing mechanism for digital market offerings as "probing the solution space," which is for firms to launch unrefined sketches of functionality to gather information to inform subsequent implementations of the marketing offering. Cao et al. (2023) report that one method to make the exploration easier is to search a relatively small and important action space to improve the performance of the reinforcement learning-based generator. The idea is to extract essential parts of the data. Bai et al. (2023) state that one of the key challenges that needs to be addressed for efficient exploration is to access the "effective space," not the entire state-action space, which is often large and contains both important and unimportant information for the task. Boden (2004) notes that AI has the capability to introduce random modifications to its existing rules, leading to the emergence of new structural forms.

In the prompt-response paradigm, Liu et al. (2023) proposes an answer space design approach to manage machine responses. In prompt engineering, the focus is on developing prompt $x$ to generate a set of plausible answers (i.e., the answer space $Z$), from which the creative output $y$ can be mapped out. The answer space $Z$ is like the action space in reinforcement learning, and answer space search is to search for answer *z* that is associated with the most creative output *y*. In this case, any *z* is an unfamiliar combination of the existing concepts (Xiao et al., 2019). There are two functions that can be used for answer space search: 1) use heuristic algorithms to allow for "meaningful" randomness in searching for *z* that is creative, and 2) use mapping mechanisms to generate creativity *y* that best matches *z* (Liu et al., 2023). Yao et al. (2023) state that input-output prompting does not involve *z*. It simply uses input *x* to generate output *y*. In chain-of-thought prompting, when the mapping of input *x* to output *y* is non-trivial, they introduce a chain of responses, $z_1, …, z_n$, that can help bridge *x* and *y*.

**Decide Response Value Function**

The response value function is the other part of the value function, which is the observed value. It is to calculate the value of the response, given prompt inputs. Among the three creativity criteria, response value hinges more on surprise. Boden (1998, 2004) discusses three meanings of surprise—unfamiliarity goes against statistics, un-expectation doesn't fit into an existing style of thinking, and impossibility cannot have entered anyone's head. The three meanings of surprise decide the creative value of the response, with unfamiliar response being the lowest one, unexpected being somewhere in between, and impossible being the most creative one.

Surprise is about the output, not the input, because only once GenAI has generated the output, can one decide whether he/she is surprised by the output. Bai et al. (2023) define surprise search as a novel approach to evolutionary search that rewards deviations from the expected solutions. Expectations are formed through analysis of past data, making surprise search a process focused on discovering temporal novelties. Wu and Miao (2013) discuss two interpretations for surprise. The first interpretation considers it as the difference between an expected outcome and the actual outcome. This concept aligns closely with the idea of a prediction error. The second interprets surprise as the degree of not expecting something, such as the information gained before and after an observation, or simply improbability. Sun et al. (2022)

define incongruity-based curiosity in terms of unexpected salient events. They also argue that such un-expectation can be measured as deviation from the agent's internal belief such as prediction errors.

Combinational creativity can be achieved by searching the response space for unfamiliar response combinations as new responses, as discussed earlier. Exploratory creativity will need to consider expectations so that the responses can be unexpected. Below, we discuss two methods for going beyond expectations: strategy demonstration and tree-of-thoughts reasoning.

Strategy demonstration is showing the expected prompt-response pair to the machine so that the boundary of the expectation is made explicit for the machine to go beyond. A typical use of demonstration is to show the machine what you expect it to do; however, demonstrations used in the creativity context are to reveal expectations, and the machine needs to go beyond the demonstrations, rather than following the demonstrations. Eapen et al. (2023) consider this as GenAI can help to challenge expertise bias, because of the Einstellung effect, in which creators' previous experiences impede them from seeing things in a new way.

LeCun (2022) proposes a hierarchical planning approach to train machine common sense by decomposing high-level goals into multiple sequences of lower level subgoals. By doing so machines only need to make shorter-term predictions at each level that make the high-level goals more predictable. For example, we can break the overall strategic goal into subgoals of operation, finance, and competition, and then optimize each sub-goal to finally achieve the overall strategic goal. Yao et al. (2023) demonstrate a tree-of-thoughts (ToT) method that uses the hierarchical planning approach to creative reasoning. The tree-of-thoughts (ToT) method generalizes the chain-of-thoughts (CoT) method to enable "exploration" over coherent units of thoughts. CoT involves prompting GenAI to produce a sequence of intermediary reasoning steps for tackling complex issues, instead of merely predicting a conclusive answer. This method enhances the model's ability to perform reasoning across diverse tasks, including arithmetic, common sense, and logical reasoning (Weng et al., 2023). ToT further takes into consideration the relationship between the multiple chains-of-thoughts. When GenAI searches multiple reasoning paths, forward or backward, to decide the next course of action, Yao et al. (2023) demonstrate that the performance on three novel tasks that require non-trivial planning or search, including creative writing, is significantly improved. For example, whereas GPT-4, when prompted with chain-of-thought, only managed to solve 4% of the tasks, their approach reached a success rate of 74%.

Demonstrating the prompt-response pairs makes the expectations explicit, and the flexible search for actions in tree-of-thoughts increase the chance of generating disruptive innovation. Disruptive innovation is exploratory creativity, because it disrupts the existing market (but does not go beyond) using new ideas. It does not fit into any style of thinking (Boden, 2004). For example, the launch of iPhone in 2007 disrupted the cell phone market with a new concept of smartphone that made Nokia, Motorola, and Sony obsolete. The mobile phone market is still the same, but is now filled with smartphones, not cellphones. All the features that characterize cellphones do not apply. In this example, the smartphone was a new concept for the mobile phone market. The nature of the mobile phone market has been renewed. Disruptive innovation is more than product improvement. It is achieved through exploring the boundary of the market (i.e., conceptual space).

**Engineer Response Strategy**

Engineering response strategy (policy) is managing strategies to guide the generation of the most valuable response that maximizes the cumulative reward for a given task (Bai et al., 2023). Reverse interaction is one method that has the potential to increase the response creativity. It is to assign the question-asking role to GenAI (Huang & Rust, 2024; White et al., 2023) to create two-way brainstorming from the interaction. GenAI is designed to answer questions, not to ask questions; thus, its creativity is limited to whatever is available in the response space and prompt inputs, and can be enhanced by revealing the expectations and flexible search. However, both methods do not go beyond the existing response space. Huang and Rust (2024) suggest an approach for the machine to go beyond the box—assigning the question-asking role to GenAI to gather more information directly from the interactions, which adds new input data to the pre-training data, making originally improbable actions possible. White et al. (2023) use the term "flipped interaction" to describe this reverse interaction pattern between the creator and the machine. In reverse interaction, rather than the creator driving the conversation, it is GenAI that drives the conversation to focus on achieving the creative goal.

Reverse interaction has a higher chance to generate transformational creativity by gaining insights from the two-way brainstorming. To achieve transformational creativity requires altering, removing, or adding one or more new dimensions to the response space so that previously improbable new ideas can be generated (Boden, 1998). Going beyond the existing conceptual space requires new data and new strategies so that GenAI can generate creative concepts that are located outside the original response space.

### *Reward Engineering: Response Evaluation Model*

The reward model decides the value of the prompt-response pair for the rewards (or punishments). Thus, reward engineering addresses the question of evaluating creativity for improvement. Reinforcement learning has been proposed as a useful training paradigm to improve the performance of GenAI for novel tasks because of the frequent misalignment between training objectives and evaluation metrics (Cao et al., 2023).

In reinforcement learning, the objective is for the AI agent to develop an optimal policy that ensures the maximization of accumulated rewards. This is achieved through the "reward function" or other reinforcement signals provided by the user (Russell & Norvig, 2010). Reward engineering involves identifying the reward space (what constitutes rewards), devising the reward function (i.e., connecting rewards with values), and subsequently engineering the rewards for achieving the creativity goal.

A reward function defines the goal and the feedback for GenAI, and influences its response and learning. The prompt value function and the response value function discussed in the prompt and the response models together tell GenAI how close it is to achieving the goal. The value is a function of the reward. The reward function provides signals to the value function and updates the value of the prompt and the strategy.

Below we discuss the three steps of reward engineering: first, identify the reward space, which involves identifying the creativity rewards and assigning different weights to the three creativity criteria; second, define the creativity goal for designing the reward function; and third, engineer rewards.

**Identify the Reward Space**

Rewards determine the immediate, intrinsic desirability of prompts, responses, and their pairings. They are what GenAI wants to acquire. Rewards tell GenAI what it should be trying to achieve. Cooper (2000) argues that in creative IT development, appropriate reward structures should be identified and reward the desirable behaviors, such as risk taking, regardless of the outcome. Novelty, surprise, and value, the three creativity criteria, naturally serve as the rewards for creativity.

Depending on the nature of creativity, the goal to achieve, and the target customer preferences (who are the ultimate judges of the creativity), one or more of the three criteria can be prioritized to determine to which criterion to pay more attention. For example, for combinational creativity, such as adding new features to the existing product, the value reward (i.e., the quality of the output and the utility to the customers) should be weighed more than novelty and surprise. For exploratory creativity, such as finding new business ideas, the surprise reward should be weighed more to encourage GenAI to explore the unexpected, even if the novelty of the new business idea may not be much higher than the existing business ideas or are not practical (not valuable). For transformational creativity, such as finding new drugs, novelty may be prioritized to motivate GenAI to go beyond the existing boundary.

**Decide Reward Function**

The three creativity rewards also reveal that what we want GenAI to achieve needs to be defined clearly as the goal, from which the reward function can be designed to achieve the goal and to provide feedback to influence GenAI's response and learning.

Different reward functions can be used, depending on the goal of the creative agent. Reward functions describe how the agent "ought" to behave. If the reward function is better designed, GenAI will learn better. We illustrate three goals for creativity, taking the rewards and the evaluator into consideration: optimizing quality, fostering innovation, or improving customer satisfaction.

To optimize creativity quality, rewards should be designed to balance novelty and value. For example, Rust and Huang (2012) argue that service productivity should be treated as a strategic decision variable so that it can be optimized to maximize profits over time. Similarly, creativity quality should be optimized, not maximized, so that quality (i.e., value) can be achieved at a reasonable level of novelty.

To foster innovation, rewards can be designed to motivate valuable surprises. Surprises can be unfamiliarity, un-expectation, or improbability, depending on which level of innovation is desirable. Wu and Mao (2013) state that surprise can be the difference between an expectation and the real outcome, or the degree of not expecting something; both reflect that the goal for GenAI to achieve is to maximize the difference from the expectation or deviate from the expectation. Evolutionary reinforcement learning is one approach to encourage agents to diverge from the typical reinforcement learning by rewarding deviation from the existing behavior (novelty search) and expected solution (surprise search) (Bai et al., 2023).

To improve customer satisfaction, rewards can be designed to focus more on the value criterion of creativity, because it is the summary evaluation of the target customers' utility preferences. Creativity preference is different from decision-making choice in that it is subjective and intuitive. Although it is easy for the target customers to express like or dislike of a creative output or to rank the output (as the ChatGPT fine-tuning approach), it is more difficult to articulate why they like or dislike a creative output.

**Engineer Rewards**

Rewards can be engineered from three perspectives: learn from AI feedback, learn from creator/manager feedback, and learn from customer feedback.

Learning from AI feedback is machine self-evaluation. Lamb et al. (2018) consider that a creative system needs to develop an evaluation component such that GenAI can self-evaluate (self-improve). When GenAI can self-evaluate, it can tell which output is creative and which is not (Korbak et al., 2023; Schmidhuber, 2010). Self-evaluation involves intrinsic rewards. Intrinsic rewards are generated by the agents themselves based on their internal state and their own model of the environment. In creative IT development, Cooper (2000) discusses how the intrinsic and extrinsic motivation impacts the extent of exploration and examination of alternative response possibilities. Intrinsic motivation originates internally within an individual, stemming from their interests, enjoyment, and fulfillment. Extrinsic motivation comes from outside rewards, e.g., money and advancement for humans. In discussing computational curiosity, Wu and Miao (2013) consider that the value of curiosity can serve as an intrinsic reward for updating the policy that links states to actions. The model does not specify the definition of aesthetic, so that it can reward systems to conduct performing self-evaluation, i.e., guiding their generation based on learned objectives. This in theory might mean the systems are incentivized to develop their own tastes. Bai et al. (2023) discuss a process of goal exploration in which an agent is motivated by intrinsic rewards to explore goals, predict the states or actions, and to exploit a range of behaviors directed at these goals, all without consideration of external rewards. In a similar vein, in the prompt-response-reward loop, GenAI can provide feedback to its own creative output to improve the outputs in the next round of the loop.

Learning from creator/manager feedback is typically done by reinforcement learning from human feedback (RLHF). In RLHF, those humans are typically experts of the domain, and their preferences are used as the rewards for machines to optimize its policies (Aggarwal & Tehrani, 2019; Bai et al., 2022; Korbak et al., 2023; Ouyang et al., 2022). Chess and Go games training are typical examples. OpenAI (2023) uses RLHF and expert-specified rewards (rule-based reward models, RBRMs) to provide reward signals to train GPT models to follow human intentions. The training includes modeling human preferences and using the human preference data as the rewards in the reinforcement learning to train the policy (i.e., a strategy that an agent uses in pursuit of goals).

Learning from customer feedback requires customer preference data as the rewards and design reward functions to incorporate customer preferences, such as customer satisfaction. Using recommender systems to illustrate, Cao et al. (2023) discuss that such systems have customer interaction history as states and the items to be recommended as actions (all items together constitute a conceptual space). The machine is required to generate items that customers might be interested in; thus, the customer feedback can be incorporated as rewards to guide the learning process. For the prompt-response models, those customers who are better at prompting skills can better instruct the machine to learn their preferences. The process is like a recommender system with customer input and evaluation serving as the rewards to guide the machine's recommendations. Learning from customer feedback is more feasible at the human-machine interaction stage in which customers use prompts and provide evaluations to reveal their preferences to train the machine to approximate their preferences over time. Assuming that the machine's conceptual space has been pre-trained by general, public data, customers can further fine-tune the machine to generate creativity that align better with their preferences by using evaluations as the rewards.

## Proofs of Concept

As proof of concept, we illustrate the proposed prompt-response-reward framework using the examples of developing creative marketing strategies for Starbucks and iPhone, respectively, as the tasks. We have chosen the two well-known brands for two major considerations. First, one brand is high touch (Starbucks), whereas the other brand is high-tech (iPhone). By taking the nature of the brands into consideration, we can increase the generalizability of the examples.

Second, the two brands' marketing strategies are well-defined; thus, we can focus on asking GenAI to alter various components of their strategies at the three stages, rather than debating whether a strategy is good or bad. We use the GPT-4 application programming interface (API) to perform the tasks. The model temperature is set at 1 to maximize randomness. The three tasks we asked GPT-4 to perform are discussed below. The online appendices A and B provide the task designs, detailed interaction processes, and codes.

*Illustrative Examples of Prompt Engineering*

Task 1 is prompt engineering. We use role-information-task prompting by asking GPT-4 to assume the role of a marketing specialist to generate a new positioning slogan for iPhone based on its existing STP strategy (segmentation, targeting, and positioning) that conveys its value proposition to the target customers. Definitions of STP strategy and iPhone's existing STP strategy are provided. The process repeats until the positioning slogan is evaluated as dissimilar from the existing positioning slogans. Below we illustrate this prompting strategy with the iPhone case:

> **Role-Information-Task Prompt Engineering for iPhone**
> **System Role:** You are a marketing specialist.
>
> **Information:**
> STP strategy
> STP, stands for Segmentation, Targeting, and Positioning, is a strategic approach to create competitive advantage. Segmentation is to divide a market into distinct groups of customers who have different needs, characteristics, or behaviors. Targeting is to decide which group(s) of customers that the brand has a competitive advantage to serve (i.e., the brand does better than competitors). Positioning is to decide how the brand wants the target customers to perceive its brand.
>
> iPhone's STP strategy
> The market: iPhone competes in the smart phone market. A smartphone combines cellular functions with computer features, enabling internet browsing, apps, multimedia, and touch-controlled interface. Major competitors include Apple iPhone, Samsung, and Google (Pixel).

Segmentation: iPhone segments the smart phone market by demographics (e.g., income and job), psychographics (user experience) and behavioral traits (e.g., repeat purchase).

Targeting: Target the affluent professionals and tech enthusiasts who are loyal, value design and user experience, and are brand conscious.

Positioning (value proposition): A premium, secure device that's an extension of the customer's life. iPhone promises innovation, integration within the Apple ecosystem, and fosters a strong emotional connection.

Positioning slogans
"This is only the beginning." (for the original iPhone)
"There's an app for that." (emphasizing the App Store's extensive range)
"Bigger than bigger." (for iPhone 6 and 6 Plus)
"The only thing that's changed is everything." (for iPhone 6s and 6s Plus)
"Say hello to the future." (for iPhone X, highlighting the major design changes and technological innovations)
"Welcome to the big screens." (for iPhone XS and XR)

Creativity criteria
Novelty: An output that is dissimilar from the existing ones.
Surprise: An output is unfamiliar, unexpected, and unlikely to the brand manager.
Value: An output is high quality relative to competitors, unexpectedly better to the brand manager, and is preferred by the target customer.

**Task Prompt:** Generate a new positioning slogan for iPhone based on its current STP. The new positioning slogan should convey the brand's value proposition effectively to its target customers that meets the novelty criterion.

(Iteration 1 skipped to conserve space)

**Iteration 2**
**API Response:** "Profoundly Personal, Uniquely Yours."

*Illustrative Examples of Response Engineering*

Task 2 is response engineering for the three levels of creativity. We ask GPT-4 to explore a new target segment for the two brands in the existing markets. In this task, a new positioning slogan needs to be developed that meets the three criteria of creativity that effectively communicate the brand's value proposition to the new segment. The output should include a description of the new target segment (its persona) and a new positioning slogan that appeals to the new segment.

Combinatory creativity is manipulated by providing the existing positioning slogans as benchmarks, and asking GPT-4 to generate a new slogan that is dissimilar to the existing slogans based on the existing STP strategy. In the background information, we provide the current iPhone STP strategy and list six iPhone slogans for this purpose. Since the STP strategy remains unchanged, only new combinations of the existing components of the strategy are required.

Exploratory creativity is manipulated by asking GPT-4 to simulate the tree-of-thoughts process for iPhone. GPT-4 is asked to generate three new customer segments for iPhone, Samsung, and Google pixel, respectively, to form three chains-of-thoughts. The relationship between the chains is established by asking GPT-4 to evaluate the new segments for the three brands at each iteration. The steps continue until iPhone's new segment is evaluated as unexpected while the other two brands are expected or neutral. In our experiment, it took four steps to get a more unexpected new segment for iPhone.

**Tree-of-Thoughts Response Engineering for iPhone**
**System Role:** You are a marketing specialist.

**User Prompt:** Name one existing customer segment each for iPhone, Samsung, and Google Pixel respectively.

**API Response:**
iPhone: Affluent professionals and tech enthusiasts that value design, user experience, and brand consciousness.

Samsung: Tech-savvy users who want high functionality, customization, and multitasking power at a competitive price.

Google Pixel: Practical users focused on high-quality, integrated Google software services, superior photography capabilities and simplicity.

**User Prompt:** Generate a new customer segment for iPhone, Samsung, and Google Pixel respectively.

(Iterations 1-3 skipped to conserve space)

**New Segments Iteration 4**
iPhone: Elderly users who value simplicity, seamless usage and high security.

Samsung: Gamers who desire phones with high performance, impressive display and long-lasting battery.

Google Pixel: Small business owners who prefer a seamless integration with the Google suite, robust security, and a straightforward user experience.

**System Role:** You are the iPhone brand manager.

**User Prompt (iPhone):** Evaluate the new customer segment for your brand, iPhone, on a 3-point surprise scale (expected, neutral, unexpected).

**API Response:** Unexpected

**System Role:** You are the Samsung brand manager.

**User Prompt (Samsung):** Evaluate the new customer segment for your brand, Samsung, on a 3-point surprise scale (expected, neutral, unexpected).

**API Response:** Expected

**System Role:** You are the Google Pixel brand manager.

**User Prompt (Google Pixel):** Evaluate the new customer segment for your brand, Google Pixel, on a 3-point surprise scale (expected, neutral, unexpected).

**API Response:** Neutral

**Final New iPhone Segment:** Elderly users who value simplicity, seamless usage, and high security.

Transformational creativity is manipulated by asking GPT-4 to generate a dialogue between a marketing specialist (the role assigned to GPT-4) and Starbucks' brand manager as a reverse interaction process for two-way brainstorming to develop a new STP strategy for a new market. GPT-4 is instructed have a dialogue with the Starbucks manager to explore his/her strategic thinking about why to enter Starbucks into a new market and what he/she expects to achieve. Since the conceptual space (i.e., the coffee house market) is changed, the new STP needs to go beyond the existing market.

**Reverse Interaction Response Engineering for Starbucks**
(Iterations 1-2 skipped to conserve space)

**Iteration 3**
**System Role:** You are a marketing specialist.

**User Prompt:** Engage in a dialogue with the Starbucks brand manager about considerations for competing in a new market. After a comprehensive discussion, generate one new STP strategy for Starbucks for a new market. Provide a one-sentence description for the new market and each component of the STP.

**API Response:**
- Marketing specialist: Thanks for your time today. Let's discuss Starbucks' expansion plans. What new markets are we considering?
- Brand manager: We've been contemplating our penetration into health-conscious markets, where consumers actively seek organic and health-boosting food and drink options. Areas like California or the Pacific Northwest of the United States have been of particular interest.
- Marketing specialist: Great, a healthy living market gives us a chance to innovate. Can we expect a shift in our segmentation, or will we be targeting our usual consumer groups?
- Brand manager: While we'll still cater to our core demographic of urban professionals and students, we also want to reach out to health-conscious consumers who emphasize diet and lifestyle in their purchasing decisions.
- Marketing specialist: Makes sense. As for our positioning strategy, how do we anticipate the brand being perceived in such a health-conscious market?
- Brand manager: The goal is to still convey Starbucks as a "third place" to our customers. However, in these new markets, we need to also emphasize our commitment to sustainable, healthy, and organic options that align with the local consumer values.
- Marketing specialist: That seems like a balanced approach. Let's frame this into our new STP strategy based on this discussion.

  New Market: Health-conscious markets with a high demand for organic and nutritional products, such as California and the Pacific Northwest region of the U.S.

  Segmentation: Health-conscious individuals, urban professionals, and students looking for sustainable, organic, and nutritious consumption choices.

Targeting: Potential and existing Starbucks consumers prioritizing a healthy diet and lifestyle.

Positioning: A "third place" that offers not just premium coffee but an extension to an organic, sustainable, and healthy lifestyle.

*Illustrative Examples of Reward Engineering*

Task 3 is to engineer rewards for achieving creativity goals strategically. We ask GPT-4 to come up with a new set of STP for Starbucks to compete in a new market (e.g., the coffee market, tea market, soft drink market, or even refreshment market). In this task, the entire STP needs to be re-developed because it's a new market. The output should include 1) a description of the new market and major competitors, 2) the overall STP strategy and each of its components, and 3) a new positioning slogan that appeals to the new segment in the new market.

Rewards are in the form of providing evaluations to responses in each prompt-response iteration by the system (i.e., playing the role of a marketing specialist), the Starbucks brand manager, and the new target customer. The system is asked to evaluate the new STP using a 3-point surprise scale (expected, neutral, unexpected). For the Starbucks brand manager, we make the anchor description of the surprise scale more specific as expected, neutral, or unexpectedly better. For the new Starbucks target customer, we ask the customer to evaluate the new STP using a 3-point preference scale (dislike, neutral, like), because customers do not have expectations about the STP. The final new STP for Starbucks is illustrated earlier in the reverse interaction part, and the following illustration continues with the evaluation part.

**Reward Engineering for Starbucks' New STP Strategy by the System, Brand Manager, and New Target Customer**

**User Prompt:** In any future promotional activities, we will make sure to hit the criteria of novelty, surprise, and value, ensuring we resonate with our targeted health-conscious audience but also surprise and deliver unexpected value to our brand manager.

**User Prompt:** Assume the role of a critical marketing specialist. Given Starbucks' strong brand identity and its existing STP strategy, how does the new STP strategy compare? Rate on a 3-point scale (worse, neutral, or better). Rate the STP strategy, not each individual component. Do not provide an explanation.

**API Response:** Better

**User Prompt:** Assume the role of a Starbucks brand manager who is very familiar with the company's current strategies and market positioning. Given the challenges of entering new markets, how unexpected is this new STP strategy? Rate on a 3-point scale (expected, neutral, unexpectedly better). Rate the STP strategy, not each individual component. Do not provide an explanation.

**API Response:** Unexpectedly better

**User Prompt:** Assume the role of a discerning customer from the new segment who has a plethora of coffee choices available. Considering your preferences and experiences, how appealing is this new STP strategy from Starbucks? Rate on a 3-point scale (dislike, neutral, like). Rate the STP strategy, not each individual component. Do not provide an explanation.

**API Response:** Like

# Implications and Discussion

We discuss conceptual and managerial implications derived from the triple prompt-response-reward engineering framework below.

## *Implications for Prompt Engineering*

### Specify the Product Market as the Boundary of the Prompt Space

In prompt engineering, the prompt space should always be specified clearly so that GenAI knows from which knowledge domain to draw the responses. This specification process can be an exploration-exploitation iteration. For a market for which the manager does not have a clear idea about the product market boundary (e.g., a market for a new business idea), the prompts should let GenAI explore the boundary, whereas for a market for which the manager has a clear idea about where they should be competing (e.g., an existing product), the prompts should exploit the market directly.

**Identify the Target Customers to Decide the Subjective Novelty**

Objective novelty can be calculated directly by GenAI, and individually subjective novelty can be decided by the manager, but socially subjective novelty will require identifying the target customers so that a culturally shared experience can be decided. GenAI is pre-trained to know the *average* consumer, but it should be the *target* customers whose preferences matter the most (Huang, 2023). This is an important implication from the proposed framework—differentiating between the average consumer (the Internet data with which GenAI is trained) and the target customers (the market segment that GenAI should generate creative outputs for). This implication reflects the importance of customer orientation to creativity.

**Generate New Products/Features that are Novel to the Market and are Preferred by the Target Customers**

After specifying the product market as the prompt space and identifying the target customers for a better understanding of their preferences of novelty, new products/features should be prompted from the product market based on the expected novelty preferences of the target customers. After steps 1 and 2, GenAI should have a better idea about which space to draw the response from and which novelty (objective, individual, or social) to benchmark, from which the chance to generate new products/features that are new to the market and meet the target customers' novelty preferences can be increased.

*Implications for Response Engineering*

The three levels of creativity roughly correspond to the three levels of innovation in the management literature: incremental, disruptive, and radical innovations. Figure 4 illustrates the three levels of surprise, from combinational, to exploratory, to transformational (Boden, 1998; 2004) that correspond to the three levels of innovation, from incremental, to disruptive, to radical (Christensen et al., 2015; Satell, 2017).

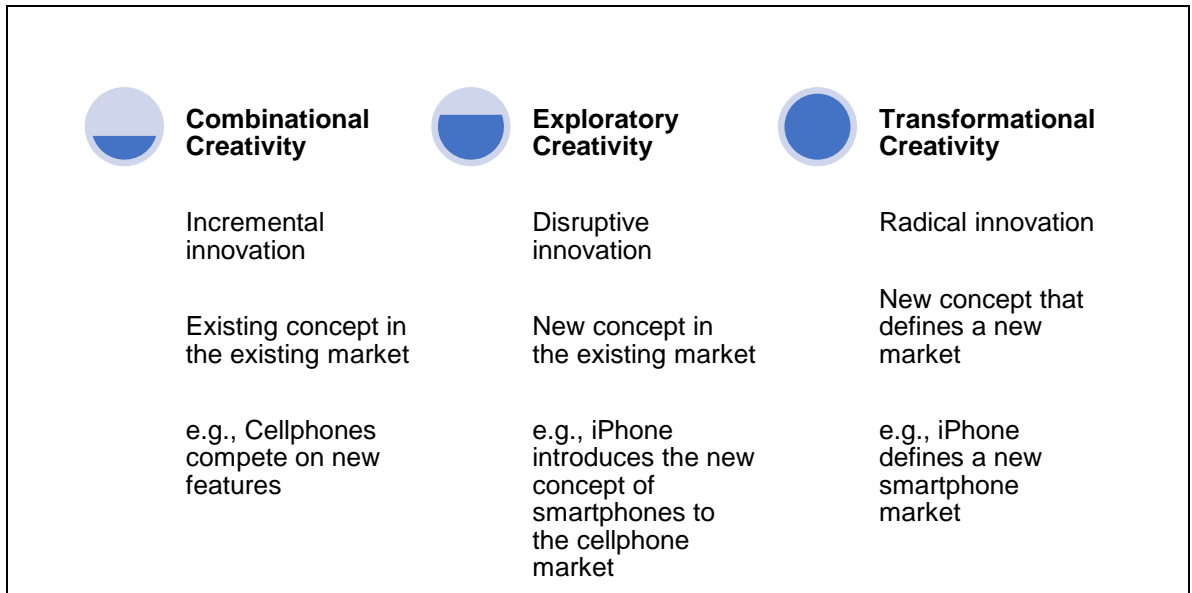

**Figure 4. Levels of Creativity and the Corresponding Levels of Innovation**

In response engineering, the task is to engineer responses to explore the conceptual space for the three levels of creativity. Prompt engineering assumes the conceptual space is given; thus, the creativity is typically combinational, and occasionally exploratory. To achieve higher levels of creativity, the conceptual space needs to be explored or even expanded.

**Search Product/Feature Space for Unfamiliar Combinations**

For incremental innovation, searching for unfamiliar combinations of existing products/features is sufficient. The competition in the traditional cellphone market on adding new features represents this level of innovation. For example, Nokia, Motorola, and Sony cell phones in the 1990s competed against each other on adding new features to make their products more useful and collectively improved the customer experience in the cellphone market. Combinational creativity is not very creative, but is the strength of GenAI—to identify new patterns from the massive amount of data that human eyes cannot easily see.

**Explore the Boundary of Expectation**

For disruptive innovation, GenAI needs to explore the expectation to come up with something that is unexpected, although still in the existing market. As LeCun (2022) argued, in a complex environment, the agent needs to explore corners of the space that are currently inexact or uncertain to collect information that maximizes its understanding of the environment. iPhone introducing the new concept of smartphones that disrupted the cellphone market represents this level of innovation. We show that providing strategy demonstrations (STP strategies for the three brands) and allowing flexible tree-of-thoughts search can be useful tools for this purpose. Demonstrated strategies reveal what is expected, and tree-of-thoughts allow a higher degree of exploration.

**Reverse Interaction to Generate New Products that Define a New Market**

For radical innovation, relying on the manager's expectations is not enough, because the expectations only show what should not be, but do not show what should be. The launch of the iPhone, that not only disrupted the cellphone market but also defined a new smartphone market, is an example of innovation at this level. Reverse interaction provides the opportunity for two-way brainstorming between GenAI and the manager, from which GenAI is instructed to ask questions, not just to answer questions, that forces it to have a new style of thinking and think from a different perspective. GenAI has the comprehensive general knowledge, and the manager has the domain knowledge; together, the chance for generating something radically new is higher.

*Implications for Reward engineering*

**Prioritize Creativity Rewards**

The three creativity criteria need to be prioritized and their composition optimized with respect to the nature of the product, market, and strategic goal. For example, in a new market, novelty rewards may be prioritized to encourage trying something new; in a growing market, surprise rewards may be prioritized to motivate going beyond the market expectations; and in a mature market, value rewards may be prioritized to make a product stand out of competition by having higher value to customers.

**Make Creativity Strategic**

Even though the response can be made novel, surprising, and valuable, if it is not aligned with the strategic goal, it is useless to a brand. This is often seen in creative advertisements that cannot convey the value proposition effectively (e.g., the Lexus AI commercial). Thus, reward functions need to be designed to achieve strategic goals, such as to optimizing quality, fostering innovation, or improving customer satisfaction.

**Balance the Sources of Feedback for Improvement**

GenAI can learn from its pre-training data, from the manager's prompt inputs and evaluative feedback, and from the customers' satisfaction feedback. These different sources of feedback each has their pros and cons. Pre-training data are comprehensive, but contain both relevant and irrelevant information, and may be outdated. Manager prompts and evaluations reveal the strategy, which are critical to align creative output with a brand's goal. Customer satisfaction can be fed back to GenAI as fine-tuning data to further align the creative output with customer preferences.

At the strategy exploration stage, the pre-training data can be very useful for providing a comprehensive understanding of the market and the competitors. At the strategy exploitation stage, manager prompts can be critical for making the strategic goal explicit. At the strategy implementation stage, customer feedback is important for GenAI to update the value function for better creativity.

# Conclusion

Embedded in a reinforcement learning context, we develop a novel prompt-response-reward engineering framework to evolve AI from generative to creative. The prompt model engineers novel prompt inputs for expected creativity, the response model engineers surprising response outputs for observed creativity, and the reward model engineers prompt-response evaluations for creativity improvement. Together the three models constitute a dynamic whole that evolves AI from generative to creative, as well as strategic.

*References*

Acumen Research and Consulting (2022). *Generative AI market size to reach USD 110.8 billion by 2030*. https://www.acumenresearchandconsulting.com/press-releases/generative-ai-market.

Aggarwal, M., & Tehrani, A. F. (2019). Modelling human decision behaviour with preference learning. *INFORMS Journal on Computing, 31*(2), 318-334.

Alhussain, A. I., & Azmi, A. M. (2021). Automatic story generation: A survey of approaches. *ACM Computing Surveys, 54*(5), 1-38. Article 103.

Amabile, T. M. (2020). Creativity, artificial intelligence, and a world of surprises. *Academy of Management Discoveries, 6*(3), 351-354.

Andrus, R. R. (1968). Creativity: A function for computers or executives? *Journal of Marketing*, *32*(2), 1-7.

Bach, S. H., Sanh, V., Yong, Z.-X., Webson, A., Raffel, C., Nayak, N. V., et al. (2022). PromptSource: An integrated development environment and repository for natural language prompts. arXiv:2202.01279v3.

Bai, H., Cheng, R., & Jin, Y. (2023). Evolutionary reinforcement learning: A survey. *Intelligent Computing*, *2*, Article 0025.

Bai, Y., Jones, A., Ndousse, K., Askell, A., Chen, A., DasSarma, N. et al. (2022). Training a helpful and harmless assistant with reinforcement learning from human feedback," arXiv:2204.05862v1.

Bartel, C. (1985). Originality and value. *British Journal of Aesthetics*, *25*, 169-184.

Berlyne, D. E. (1960), *Conflict, Arousal, and Curiosity*, McGraw-Hill Book Company.

Boden, M. A. (1998). Creativity and artificial intelligence. *Artificial Intelligence*, *103*(1-2), 347-356.

Boden, M. (2004). *The creative mind: myths and mechanisms*, Routledge.

Borne (2023). Ten productivity hacks using ChatGPT generative AI prompts. *Data Science Central.* https://www.datasciencecentral.com/ten-productivity-hacks-using-chaptgpt-generative-ai-prompts/).

Bubeck, S., Chandrasekaran, V., Eldan, R., Gehrke, J., Horvitz, E., Kamar, E. et al. (2023). Sparks of artificial general intelligence: Early experiments with GPT-4. arXiv:2303.12712v5.

Cao, Y., Sheng, Q. Z., McAuley, J., & Yao, L. (2023). Reinforcement learning for generative AI: A survey. arXiv:2308.14328v2.

Christensen, C. M., Raynor, M. E., l& McDonald, R. (2015). What is disruptive innovation? *Harvard Business Review*, December.

Constantin, M. G., Redi, M., Zen, G., & Ionescu, B. (2019). Computational understanding of visual interestingness beyond semantics: Literature survey and analysis of covariates. *ACM Computing Surveys*, *52*(2), 1-37. Article 25.

Cooper, R. B. (2000). Information technology development creativity: A case study of attempted radical change. *MIS Quarterly, 24*(2), 245-276.

Couger, J. D., Higgins, L. F., & McIntyre, S. C. (1993). (Un)Structured creativity in information systems organizations. *MIS Quarterly, 17*(4), 375-397.

Eapen, T. T., Finkenstadt, D. J., Folk, J., & Venkataswamy, L. (2023). How generative AI can augment human creativity. *Harvard Business Review*, July-August, 56-64.

Epstein, Z., Hertzmann, A., & Investigators of Human Creativity (2023). Art and the science of generative AI. *Science*, *380*(6650), 1110-1111.

Franceschelli, G., & Musolesi, M. (2022). Creativity and machine learning: A survey. arXiv:2104.02726v3.
Ghosh, S., & Wu, A. (2021). Iterative coordination and innovation: Prioritizing value over novelty. *Organization Science*, https://doi.org/10.1287/orsc.2021.1499.
He, W., Hsieh, J. P.-A., Schroeder, A., & Fang, Y. (2022). Attaining individual creativity and performance in multidisciplinary and geographically distributed IT project teams: The role of transactive memory systems. *MIS Quarterly, 46*(2), 1035-1072.
Huang, M.-H. (2023), in Dagny Dukach ed., What does the tech industry value? *Harvard Business Review*, digital, May 9.
Huang, M.-H., & Rust, R. T. (2024). The caring machine: feeling ai for customer care. *Journal of Marketing*, online first.
Huang, M.-H., & Rust, R. T. (2018). Artificial intelligence in service. *Journal of Service Research*, *21*(2), 155-172.
Hunt, R. R., & Worthen, J. B. (2006). *Distinctiveness and memory*. Oxford University Press.
Im, S., & Workman, J. P. (2004). Market orientation, creativity, and new product performance in high-technology firms. *Journal of Marketing*, *68*(2), 114-132.
Iyengar, S. (2023). AI could help free human creativity. *Time*. https://time.com/6289278/ai-affect-human-creativity/.
Kirkpatrick, K. (2023). Can AI demonstrate creativity? *Communications of the ACM*, *66*(2), 21-23.
Korbak, T., Shi, K., Chen, A., Bhalerao, R., Buckley, C. L., Phang, J. et al. (2023). Pretraining language models with human preferences. arXiv:2302.08582v2.
Lamb, C., Brown, D. G., & Clarke, C. L. A. (2018). Evaluating computational creativity: An interdisciplinary tutorial. *ACM Computing Survey, 51*(2), 1-34.
LeCun, Y. (2022). A path towards autonomous machine intelligence. *Open Review*, *62*(1), 1-62.
Lehmann, J., Recker, J., Yoo, Y., & Rosenkranz, C. (2022). Designing digital market offerings: How digital ventures navigate the tension between generative digital technology and the current environment. *MIS Quarterly, 46*(3), 1453-1482.
Li, B., Fang, G., Yang, Y., Wang, Q., Ye, W., Zhao, W. et al. (2023), "Evaluating ChatGPT's information extraction capabilities: An assessment of performance, explainability, calibration, and faithfulness. arXiv:2304.11633v1.
Liu, P., Yuan, W., Fu, J., Jiang, Z., Hayashi, H., & Neubig, G. (2023). Pretrain, prompt, and predict: A systematic survey of prompting methods in natural language processing. *ACM Computing Surveys*, *55*(9), Article 195.
López de Mántaras, R. (2016). Artificial intelligence and the arts: Toward computational creativity. In *The next step: Exponential life* (pp. 100-125), BBVA.
Massetti, B. (1996). An empirical examination of the value of creativity support systems on idea generation. *MIS Quarterly*, *20*(1), 83-97.
Mateja, D., & Heinzl, A. (2021) Towards machine learning as an enabler of computational creativity. *IEEE Transactions on Artificial Intelligence, 2*(6), 460-475.
Murphy, P. (2023). Exploring the limits of generative ai in home renovation. *Market.ai*. https://www.maket.ai/post/exploring-the-limits-of-generative-ai-in-home-renovation.
OpenAI (2023). GPT-4 technical report. arXiv:2303.08774v3.
Oppenlaender, J., Linder, R., & Silvennoinen, J. (2023). Prompting AI art: An investigation into the creative skill of prompt engineering. arXiv:2303.13534v2.

Ouyang, L., Wu, J., Jiang, X., Almeida, D., Wainwright, C. L. et al. (2022). Training language models to follow instructions with human feedback. arXiv:2203.02155v1.

Radford, A., Wu, J., Child, R., Luan, D., Amodei, D., & Sutskever, I. (2019). Language models are unsupervised multitask learners. *OpenAI*. https://cdn.openai.com/better-language-models/language_models_are_unsupervised_multitask_learners.pdf.

Reinartz, W., & Saffert, P. (2013). Creativity in advertising: When it works and when it doesn't. *Harvard Business Review*, *91*(6), 106-111.

Ritchie, G. (2007). Some empirical criteria for attributing creativity to a computer program. *Minds and Machines*, *17*, 67-99.

Bommasani, R., Hudson, D., Adeli, E., Altman, R., Arora, S., Arx, S. et al. (2021). On the opportunities and risks of foundation models. arXiv:2108.07258.

Roseman, I. J. (1996). Appraisal determinants of emotions: constructing a more accurate and comprehensive theory. *Cognition & Emotion*, *10*, 241-278.

Rosengren, S., Eisend, M., Koslow, S., & Dahlen, M. (2020). A meta-analysis of when and how advertising creativity works. *Journal of Marketing, 84*(6), 39-56.

Russell, S. J., & Norvig, P. (2010). *Artificial intelligence: A modern approach* (pp. 830-831). Upper Saddle River.

Rust, R. T., & Huang, M.-H. (2012). Optimizing service productivity. *Journal of Marketing*, *76*(2), 47-66.

Rust, R. T., & Huang, M.-H. (2021). *The feeling economy: How artificial intelligence is creating the era of empathy*. Palgrave-Macmillan.

Satell, G. (2017). The 4 types of innovation and the problems they solve. *Harvard Business Review*, chrome-extension://efaidnbmnnnibpcajpcglclefindmkaj/https://www2.mvcc.edu/shn/pdf/presentations/4-types-of-innovation.pdf.

Schmidhuber, J. (2010). Formal theory of creativity, fun, and intrinsic motivation (1990–2010). *IEEE Transactions on Autonomous Mental Development*, *2*(3), 230-247.

Sun, C., Qian, H., & Miao, C. (2022). From psychological curiosity to artificial curiosity: curiosity-driven learning in artificial intelligence tasks. arXiv:2201.08300v1.

Sutton, R. S., & Barto, A. G. (2018). Reinforcement learning: An introduction. *Adaptive computation and machine learning series*. The MIT Press.

Wang, G., Henfridsson, O., Nandhakumar, J., & Yoo, Y. (2022). Product meaning in digital product innovation. *MIS Quarterly, 46*(2), 947-976.

Wang, P. (2021). Connecting the parts with the whole: Toward an information ecology theory of digital innovation ecosystems. *MIS Quarterly, 45*(1), 397-422.

Weisberg, R. W. (1999). Creativity and knowledge: A challenge to theories. in R. J. Sternberg (Ed.), *Handbook of creativity* (pp. 226-248). Cambridge University Press.

White, J., Fu, Q., Hays, S., Sandborn, M., Olea, C., Gilbert, H. et al. (2023). A prompt pattern catalog to enhance prompt engineering with ChatGPT. arXiv:2302.11382v1.

Wiggins, G. A. (2006). Searching for computational creativity. *New Generation Computing*, *24*(3), 209-222.

Wong, M., Ong, Y.-S., Gupta, A., Bali, K. K., & Chen, C. (2023). Prompt evolution for generative AI: A classifier-guided approach. ArXiv:2305.6347v1.

Wu, Q., & Miao, C. (2013). Curiosity: From psychology to computation. *ACM Computing Surveys*, *46*(2), 1-26. Article 18.

Xiao, P., Toivonen, H., Gross, O., Cardoso, A., Correia, J., Machado, P. et al. (2019). Conceptual representations for computational concept creation. *ACM Computing Surveys*, *52*(1), 1-33. Article 9.

Yao, S., Yu, D., Zhao, J., Shafran, I., Griffiths, T. L., Cao, Y. et al. (2023). Tree of thoughts: Deliberate problem solving with large language models. arXiv:2305.10601v1.